\documentclass{article}

\usepackage{PRIMEarxiv}

\usepackage[utf8]{inputenc} % allow utf-8 input
\usepackage[T1]{fontenc}    % use 8-bit T1 fonts
\usepackage{hyperref}       % hyperlinks
\usepackage{url}            % simple URL typesetting
\usepackage{booktabs}       % professional-quality tables
\usepackage{amsfonts}       % blackboard math symbols
\usepackage{nicefrac}       % compact symbols for 1/2, etc.
\usepackage{microtype}      % microtypography
\usepackage{lipsum}
\usepackage{fancyhdr}       % header
\usepackage{graphicx}       % graphics
\usepackage{amsmath,amssymb,amsfonts}
\usepackage{makecell}
\usepackage{multirow}
\graphicspath{{media/}}     % organize your images and other figures under media/ folder

\title{ContrastCAD: Contrastive Learning-based Representation Learning for Computer-Aided Design Models
}

\author{
  Minseop Jung, Jibum Kim \\
  Department of Computer Science and Engineering \\
  Incheon National University \\
  Incheon, South Korea\\
  \texttt{\{minseob22, jibumkim\}@inu.ac.kr} \\
   \And
  Minseong Kim \\
  CareSquare \\
  Seoul\\
  \texttt{qtddpms@gmail.com} \\
   \And
}

\begin{document}
\maketitle

\begin{abstract}
The success of Transformer-based models has encouraged many researchers to learn CAD models using sequence-based approaches. However, learning CAD models is still a challenge, because they can be represented as complex shapes with long construction sequences. Furthermore, the same CAD model can be expressed using different CAD construction sequences. We propose a novel contrastive learning-based approach, named ContrastCAD, that effectively captures semantic information within the construction sequences of the CAD model. ContrastCAD generates augmented views using dropout techniques without altering the shape of the CAD model. We also propose a new CAD data augmentation method, called a Random Replace and Extrude (RRE) method, to enhance the learning performance of the model when training an imbalanced training CAD dataset. Experimental results show that the proposed RRE augmentation method significantly enhances the learning performance of Transformer-based autoencoders, even for complex CAD models having very long construction sequences. The proposed ContrastCAD model is shown to be robust to permutation changes of construction sequences and performs better representation learning by generating representation spaces where similar CAD models are more closely clustered. Our codes are available at \url{https://github.com/cm8908/ContrastCAD}.
\end{abstract}

% keywords can be removed

\section{Introduction}\label{sec:introduction}

CAD models are widely used in various industrial applications such as product design, circuit design, and component manufacturing. They are used from initial design to final product production when designing 3D products. These CAD models are often generated by experienced CAD designers using commercial CAD tools such as Solidworks and AutoCAD. However, generating complex and professional CAD models is a challenge, even when using well-known CAD tools, because CAD models involve various geometric shapes and details, requiring professional CAD expertise. CAD models are represented in the form of sequences with multiple operations during the design process in CAD tools, known as \textit{construction sequences}, and each operation represents a drawing step of the CAD model, which finally results in the 3D shape of the CAD model. 

CAD models are basically 3D shapes, and many studies have been conducted in the computer vision domain. There have been many attempts to learn CAD models in a discretized form such as point clouds or polygonal meshes \cite{point2geo, parsenet, pienet, inferring-cad}. However, these discretized forms have limitations in learning CAD models as they neglect key shape details that CAD designers consider when creating CAD models. CAD designers carefully design CAD models in sequence using construction sequences, but these aspects are ignored when we learn CAD models using discretized representations. In contrast, numerous researchers have tried to learn CAD models using construction sequences owing to the similarity between sequences in natural language processing and construction sequences in the CAD model. In particular, Transformer-based models that are highly successful in natural language processing and computer vision areas exhibited excellent performance for learning CAD models based on construction sequences \cite{deepcad, skexgen, hiernet}. One of the pioneering works for learning CAD models using Transformer-based models is DeepCAD \cite{deepcad}, which represents CAD models in the form of \textit{sketch} and \textit{extrusion} pairs using construction sequences and reconstructs CAD models using a Transformer-based autoencoder. 
 
Although recent Transformer-based models have demonstrated some success in learning CAD models, there are still major issues that need resolution. One of the major difficulties is that a single CAD model can be represented by multiple CAD construction sequences. Figure \ref{fig:fig1}(a) shows an example where two different CAD construction sequences generate the same CAD models. Furthermore, similar construction sequences can generate very different CAD models, as shown in Figure \ref{fig:fig1}(b). This example also demonstrates that a few operation changes in construction sequences can produce completely different CAD models. This explains the need for employing contrastive learning when learning CAD models: the model should learn to position similar CAD data close to each other in the latent space within the training dataset, while dissimilar CAD data should be positioned farther apart.
 
Various CAD data augmentation methods have been studied by researchers to enhance the learning performance of the model. DeepCAD, a well-known CAD training dataset, is an imbalanced dataset where the proportions of each command within the construction sequence are not uniform \cite{deepcad}. Therefore, when a model is trained on this imbalanced dataset, it cannot sufficiently learn some 3D shapes based on specific commands; consequently, the diversity of the learned CAD model may be insufficient. We also point out that many previous works have encountered considerable difficulties in learning long construction sequences in the training dataset that are necessary for representing complex CAD models. Various attempts have been made to address these issues using data augmentation techniques to increase the training data so as to enhance the learning performance. Wu et al. proposed a method that swaps some operations between two different construction sequences in the training dataset \cite{deepcad}. While their CAD data augmentation method is effective in increasing CAD data and also improves the learning performance of the model by generating completely new 3D shapes in the training dataset, it is  applicable to only a subset of the training data. Xu et al. augmented training data by adding random noise to the coordinates of geometry tokens \cite{skexgen}. However, their approach only adds some noise to existing shapes and has limitations in terms of generating diverse and entirely new 3D CAD shapes. 

We propose a novel CAD model learning approach based on contrastive learning. The proposed CAD model learning comprises three main aspects. First, we propose a new CAD data augmentation method, called Random Replace and Extrude (RRE), to enhance the learning performance of the model when training an imbalanced training CAD dataset. Second, we effectively reconstruct CAD construction sequences using a Transformer-based autoencoder. The proposed Transformer-based autoencoder takes construction sequences as input, performs embedding and self-attention in the encoder to create latent vectors, and reconstructs construction sequences from these latent vectors in the decoder. 

As a final step, we propose a ContrastCAD model based on contrastive learning to effectively capture semantic information within the construction sequences of the CAD model. Contrastive learning aims to train positive pair data to have similar embedding values while ensuring that negative pair data have dissimilar embedding values. In computer vision communities, transformation-based contrastive learning is widely used because various transformation-based techniques are simple and effective techniques for image-based applications, but these techniques can lead to invalid CAD models when applied to CAD models having 3D shapes. 

We are inspired by a dropout technique in natural language processing that masks a certain portion of embedding values to zero randomly and considers the resulting embedding vectors as positive pairs. We apply two different dropout masks to the latent vectors to form positive pairs and construct negative pairs from latent vectors of different CAD models within the same batch. In this way, we generate augmented views without changing the shape of the CAD model. Our goal is to train latent vectors of similar CAD models to be closer and latent vectors of dissimilar CAD models to be further apart. We also aim to show that our model's representation space, through contrastive learning, exhibits better clustering results compared to existing approaches. Finally, when the proposed ContrastCAD is well-trained, we can automatically generate diverse and complex CAD models from learned latent vectors.

Our contributions can be summarized as follows:

\begin{itemize}
\item We propose the ContrastCAD model, which performs novel representation learning by augmenting the embedding for CAD models through contrastive learning.
\item The proposed ContrastCAD trains similar CAD models to be closer in the latent space by reflecting the semantic information of construction sequences more effectively. ContrastCAD is also robust to the permutation changes of CAD construction sequences. 
\item We introduce a new CAD data augmentation method, RRE data augmentation, which can be applied to all CAD training data and substantially improves the accuracy of the autoencoder during reconstruction. 
\end{itemize}

\begin{figure}[!t]
\centering    
\includegraphics[width=\columnwidth]{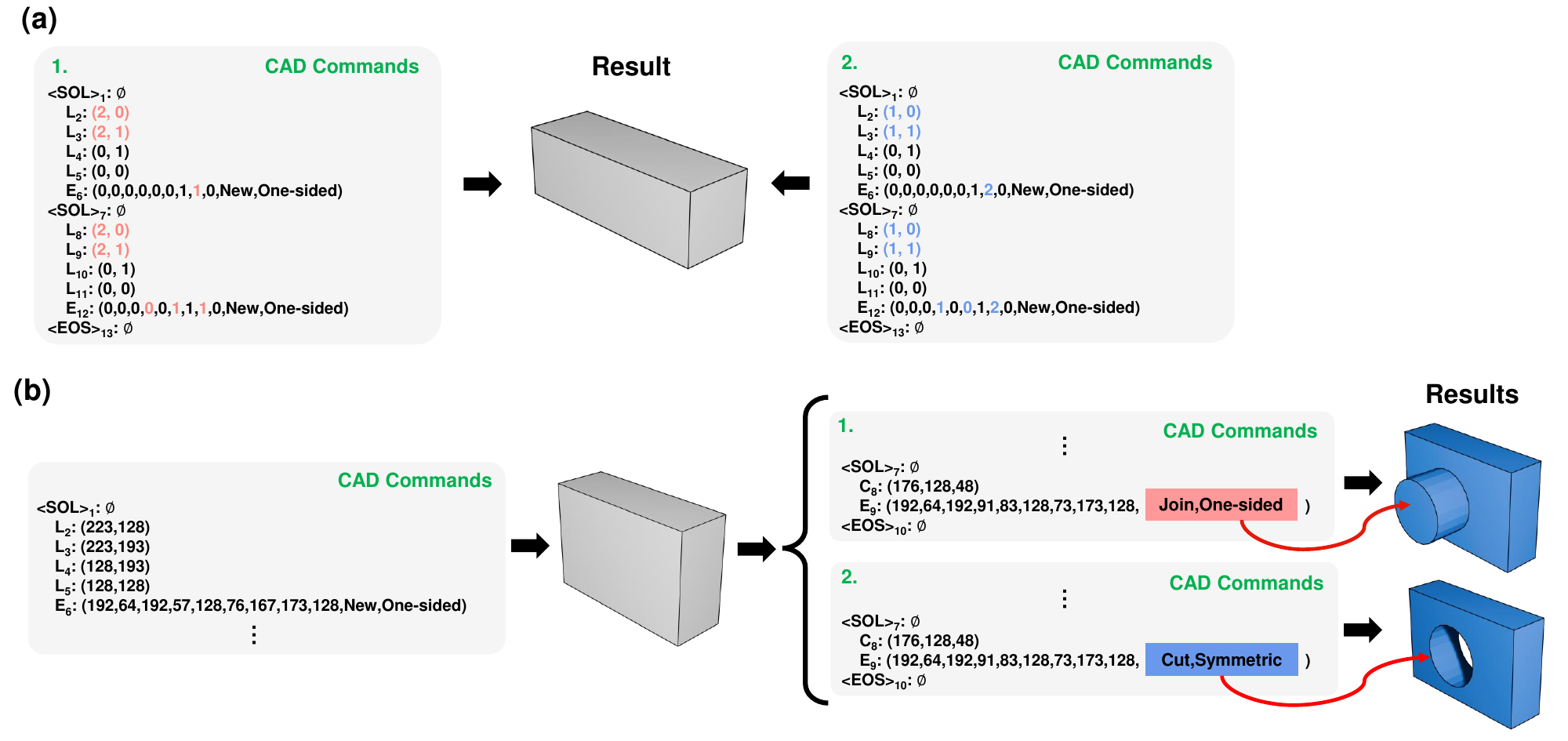}
   \caption{Two examples illustrating the difficulties of CAD model training. (a) Two different CAD construction sequences can generate the same CAD model and (b) a few operation changes in the construction sequence produce a completely different CAD model.} 
   \label{fig:fig1}
\end{figure}

%%%%%%

\section{Related works}
\label{sec:related works}
\subsection{3D representation learning}

For learning representation of 3D models, researchers train neural networks using data structures such as point clouds, meshes, and voxels to discretize 3D shapes. Researchers have also proposed approaches to extract geometric features from point clouds to construct 3D surfaces \cite{point2geo, parsenet, pienet}. Xu et al. represented boundary representation (B-rep) from mesh and point cloud data, introducing a new geometric representation called zone graph from neural networks \cite{inferring-cad}. Lambourne et al. proposed representing CAD models from voxel data using signed distance functions \cite{voxel2cad}. Mo et al. successfully represented the hierarchical features of 3D shapes using graph neural networks \cite{structurenet}. Jones et al. synthesized 3D shapes successfully based on geometric representations \cite{shapeassembly}. Other researchers efficiently represented geometric features of CAD models represented by B-Rep using neural networks \cite{uvnet, brepnet, complexgen}.

\subsection{3D CAD Generation}
There has been considerable progress in automatically generating CAD models using generative models. Researchers successfully generated CAD models by representing the Boolean operations of CAD models expressed in constructive solid geometry as hierarchical trees \cite{csgnet, ucsgnet, d2csg}. Lambourne et al. proposed a method to generate CAD models from voxel data using neural networks \cite{voxel2cad}. Researchers suggested methods to represent and automatically generate CAD models from 2D hand-drawn sketches \cite{sketch2cad, vitruvion, free2cad, cad-sketches}. However, these approaches require additional steps to convert the generated CAD models into a format usable in CAD tools for editing. 

 Consequently, recent studies have attempted to represent CAD models as construction sequences that are user-editable. Wu et al. proposed a method to represent CAD models belonging to \cite{abcdataset} as construction sequences that are composed of CAD commands and trained them using a Transformer-based autoencoder \cite{deepcad}. Other researchers proposed an approach to generate CAD models using construction sequences that are represented as discrete codebooks \cite{skexgen, hiernet}. CAD models represented by these construction sequences are user-editable, and Transformer-based models that are successfully used in the field of natural language processing can be used to learn these construction sequences. 

\subsection{Contrastive learning}
Supervised learning-based approaches require a large amount of labeled data. There has been considerable research on self-supervised learning. In particular, contrastive-learning-based approaches have received immense attention with a view to conduct representation learning for image or text data without labeled data. Researchers improved the recognition performance of neural networks for images through contrastive learning, using data obtained through different data augmentation methods for a single image \cite{simclr, simclr-v2}. Other researchers effectively mitigated representation collapse during training by handling negative pairs and positive pairs used in contrastive learning \cite{moco, byol, siamsiam}. In the field of natural language processing, Gao et al. demonstrated that contrastive learning is possible without data transformation by applying dropout techniques, thereby alleviating representation collapse issues \cite{simcse}. Chuang et al. learned natural language sentence embeddings through a difference-based approach that performs contrastive learning between natural language sentences and their masked counterparts \cite{diffcse}. 

 Recently, Ma et al. represented CAD models in a multimodal fashion and introduced dropout-based contrastive learning similar to \cite{simcse} for effective representation learning across different modalities \cite{multicad}. While this approach is effective for learning multimodal representations, the proposed contrastive learning method has limitations in that it requires the existence of data with at least two modalities and also requires additional steps to align different modalities. 

%%%%%%

\section{CAD Construction Sequence}
\label{sec:cad construction sequence}

% Table 1
\begin{table}[]
    \centering
    \caption {CAD commands and parameters.}
    \label{tab:tab1}
    \resizebox{.7\columnwidth}{!}{%
    \begin{tabular}{lll}
    \Xhline{2\arrayrulewidth}
    Commands                  & Parameters          & Description                     \\ 
    \Xhline{2\arrayrulewidth}
    $\langle\text{SOL}\rangle$                          & $\emptyset$                 &                                 \\ \hline
    $L$ (Line)                       & $x$, $y$                    & line end point                  \\ \hline
    \multirow{3}{*}{A (Arc)}       & $x$, $y$                    & arc end point                   \\
                                   & $\theta$                    & sweep angle                     \\
                                   & $c$                         & counter-clockwise flag                  \\ \hline
    \multirow{2}{*}{C (Circle)}    & $x$, $y$                    & circle center point             \\
                                   & $r$                         & circle radius                   \\ \hline
    \multirow{6}{*}{E (Extrusion)} & $\alpha$, $\beta$, $\gamma$ & sketch plane orientation        \\
                                   & $o_x$, $o_y$, $o_z$         & sketch plane origin             \\
                                   & $s$                         & scale of the sketch profile     \\
                                   & $\delta_1$, $\delta_2$      & extrude distance for both sides \\
                                   & $b$                         & boolean operation               \\
                                   & $w$                         & extrude type                    \\ \hline
    $\langle\text{EOS}\rangle$                          & $\emptyset$                 &                                 \\ \Xhline{2\arrayrulewidth}
    \end{tabular}%
    }
    \end{table}

We represent and learn CAD models using construction sequences that are composed of a series of CAD commands. Each construction sequence is in the form of a sequence of commands, making it easily understandable for humans and exportable to commercial CAD tools such as Solidworks and AutoCAD. In addition, it can be easily transformed into other forms for representing CAD models, such as B-rep. Recent research has successfully used construction sequences to learn CAD models \cite{deepcad, skexgen, hiernet}.

The commands in the construction sequence can be divided into two parts: \textit{sketch} and \textit{extrusion} commands. The sketch commands draw closed curves on a 2D plane, referred to as \textit{loops}. Each loop begins with the $\langle\text{SOL}\rangle$ command, followed by a series of commands, $C_i$, sequentially. There are three types of commands representing sketches: \textit{line}, \textit{arc}, and \textit{circle}. The extrusion command extends 2D sketch profiles from a 2D plane into a 3D body. 

The extrusion command includes parameters that define the orientation and origin of the 2D sketch plane, as well as the extrusion type and the merging form with the existing 3D body. The extrusion type can be one-sided, symmetric, or two-sided, depending on the sketch plane of the profile. When the generated 3D body is combined with a previously created body, the merging with the previously created shape during extrusion determines whether the existing 3D body will be joined, cut, or intersected. The extrusion command includes parameters for the scale of the associated sketch profile and the extrusion distance. The commands and parameters used in the paper are listed in Table \ref{tab:tab1}, comprising a total of 16 types of parameters. Readers can refer to \cite{deepcad} for further details of each parameter.

\begin{figure}[!t]
\centering    
\includegraphics[width=.9\columnwidth]{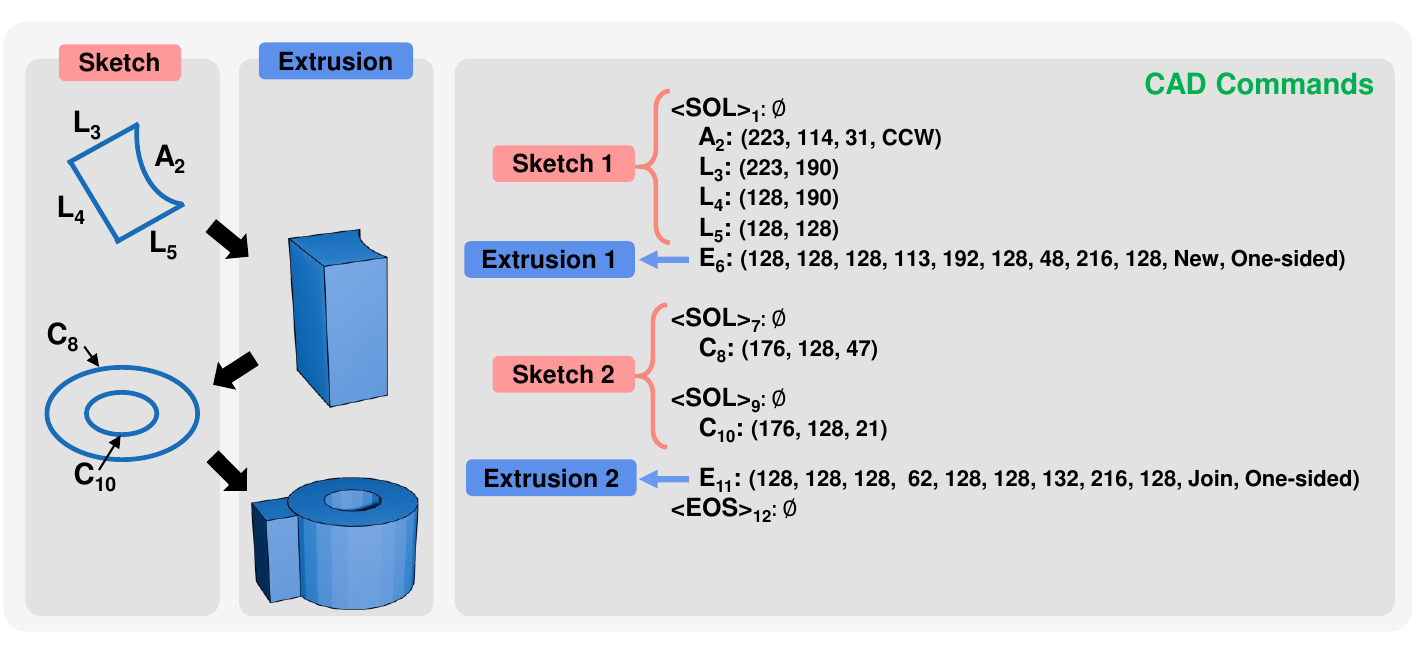}
   \caption{CAD model example defined by two sketch commands and two extrusion commands. In “Sketch 1”, four consecutive commands ($\text{A}_2$, $\text{L}_3$, $\text{L}_4$, $\text{L}_5$) form a loop and the extrusion command $\text{E}_6$ generates a single 3D body in Extrusion 1. In “Sketch 2”, two consecutive commands ($\text{C}_8$ and $\text{C}_{10}$) form two loops and the extrusion command, $\text{E}_{11}$, generates another 3D body. It merges with the previously created 3D body in a joining form.} 
   \label{fig:fig2}
\end{figure}

Figure \ref{fig:fig2} illustrates a CAD model example defined by sketch and extrusion commands. In this example, “Sketch 1” was defined using four consecutive commands ($\text{A}_2$, $\text{L}_3$, $\text{L}_4$, $\text{L}_5$), followed by using the extrusion command $\text{E}_6$ to generate a single 3D body. The extrusion type was one-sided, as extrusion was performed in one direction. Then, “Sketch 2” was created using two consecutive commands ($\text{C}_8$ and $\text{C}_{10}$), followed by using the extrusion command $\text{E}_{11}$ to generate another 3D body. Next, one-sided extrusion was performed. Finally, during extrusion, it was merged with the previously created 3D body in a joining form.

%%%%%%%%%%%%

\section{Method}
\label{sec:method}
\begin{figure}[!t]
\centering    
\includegraphics[width=\columnwidth]{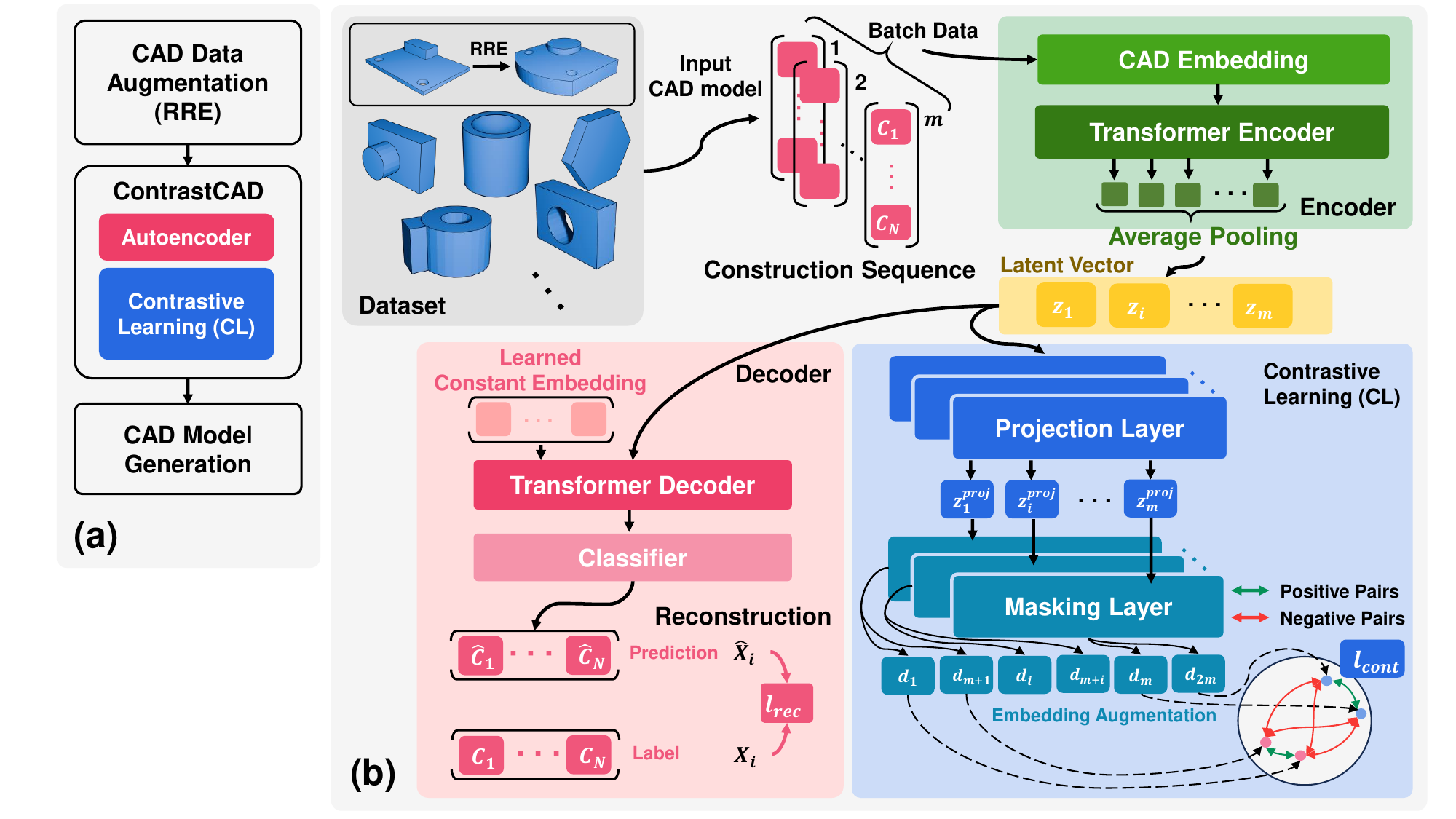}
   \caption{(a) Overview of the proposed CAD model learning and generation method and (b) proposed ContrastCAD model based on contrastive learning.} 
   \label{fig:fig3}
\end{figure}

The proposed CAD model learning and generation method are illustrated in Figure \ref{fig:fig3}(a). First, the proposed RRE CAD data augmentation method augments CAD training data to enhance the training capability of the model. Second, we propose a new contrastive learning-based model, named ContrastCAD, as shown in Figure \ref{fig:fig3}(b). In ContrastCAD, a Transformer-based autoencoder is employed to efficiently reconstruct the CAD construction sequences and learn the latent vectors of the construction sequences based on contrastive learning. Finally, once ContrastCAD is well-trained, CAD models are automatically generated using the learned latent vectors. 

\subsection{RRE CAD Data Augmentation}

% Table 2
\begin{table}[]
\centering
\caption {Number and proportion of construction sequences with respect to three command types in the sketch command: line, circle, and arc.}
\label{tab:tab2}
\resizebox{.7\columnwidth}{!}{%
\begin{tabular}{llll}
\Xhline{2\arrayrulewidth}
\begin{tabular}[c]{@{}l@{}}\# total \\ sequences \\ (proportion)\end{tabular} & \begin{tabular}[c]{@{}l@{}}\# sequences \\ including lines \\ (proportion)\end{tabular} & \begin{tabular}[c]{@{}l@{}}\# sequences \\ including circles \\ (proportion)\end{tabular} & \begin{tabular}[c]{@{}l@{}}\# sequences\\ including arcs \\ (proportion)\end{tabular} \\ 
\Xhline{2\arrayrulewidth}
\begin{tabular}[c]{@{}l@{}}179,133 \\ (100\%)\end{tabular}
& \begin{tabular}[c]{@{}l@{}}140,406 \\ (78.38\%)\end{tabular}                              & \begin{tabular}[c]{@{}l@{}}76,694 \\ (42.81\%)\end{tabular}                                 & \begin{tabular}[c]{@{}l@{}}35,392 \\ (19.76\%)\end{tabular}                             \\ \Xhline{2\arrayrulewidth}
\end{tabular}%
}
\end{table}

% Table 3
\begin{table}[]
\centering
\caption {Number and proportion of construction sequences with respect to three extrusion types: one-sided, symmetric, two-sided. }
\label{tab:tab3b}
\resizebox{.7\columnwidth}{!}{%
\begin{tabular}{llll}
\Xhline{2\arrayrulewidth}
\begin{tabular}[c]{@{}l@{}}\# total sequences\\ (proportion)\end{tabular} & \begin{tabular}[c]{@{}l@{}}\# sequences \\  $w$=One-sided\\ (proportion)\end{tabular} & \begin{tabular}[c]{@{}l@{}}\# sequences \\  $w$=Symmetric\\ (proportion)\end{tabular} & \begin{tabular}[c]{@{}l@{}}\# sequences \\ $w$=Two-sided\\ (proportion)\end{tabular} \\ 
\Xhline{2\arrayrulewidth}
\begin{tabular}[c]{@{}l@{}}179,133\\ (100\%)\end{tabular}                                      & \begin{tabular}[c]{@{}l@{}}165,876 \\ (92.60\%)\end{tabular}                                  & \begin{tabular}[c]{@{}l@{}}16,126 \\ (9.00\%)\end{tabular}                                   & \begin{tabular}[c]{@{}l@{}}3,059  \\ (1.71\%)\end{tabular}                                   \\ 
\Xhline{2\arrayrulewidth}
\end{tabular}%
}
\end{table}
The DeepCAD dataset \cite{deepcad} we utilized as our training dataset is a large-scale dataset representing 3D CAD models as construction sequences, commonly employed in various previous studies for training 3D CAD models. However, this dataset suffers from the following issues.

First, there are three types of commands that represent sketches: line, arc, and circle. However, the proportion of construction sequences containing each sketch command in the dataset is not uniform. Table \ref{tab:tab2} presents the proportion of sequences containing each sketch command in the DeepCAD dataset. As presented in the table, 78.38\% of the sequences contain the line command, whereas only 19.76\% contain the arc command. Consequently, the model may suffer from limitations in reconstruction performance for CAD models that include such sketch commands.

Second, the extrusion type parameter of the extrusion command can have three type of values: one-sided, symmetric, and two-sided. However, most CAD models in the dataset are labeled as one-sided, leading to a limitation in the diversity of 3D shapes created through extrusion. Table \ref{tab:tab3b} presents the proportion of sequences with extrusion types labeled as one-sided, symmetric, and two-sided in the DeepCAD dataset. As listed in the table, sequences with the one-sided extrusion type account for 92.60\%, while those with symmetric and two-sided extrusion types account for only 9.00\% and 1.71\%, respectively.

We propose a new method that augments CAD training data called the Random Replace and Extrude (RRE) method. The proposed RRE data augmentation method operates as follows. First, a portion of the line commands in the construction sequences from the dataset is randomly selected and replaced with arc commands. The parameters of the replaced arc commands, namely the $x$ and $y$ parameters, remain the same as those of the original line commands, while the sweep angle ($\theta$) parameter is randomly sampled from a uniform distribution from the integers in the range $\left[1, 255\right]$, and the counter-clockwise flag ($c$) parameter is randomly chosen between 0 and 1. This allows the model to better learn arc commands that need to be augmented more in the training dataset and introduces various new 3D shapes by transforming straight lines into curves. 

Second, the extrusion type ($w$ in Table \ref{tab:tab1}) and extrusion distance parameters ($\delta_1$, $\delta_2$ in Table \ref{tab:tab1}) of the extrusion commands are randomly changed. Specifically, for $w$, values from the set $\{0, 1, 2\}$ (representing One-sided, Symmetric, Two-sided, respectively) are randomly sampled from a uniform distribution, while $\delta_1$ and $\delta_2$ are randomly sampled from a uniform distribution from integers in the range $\left[0, 255\right]$. This enables the model to learn a wider variety of 3D shapes produced through extrusion, including those with symmetric or two-sided extrusion types. As a final step, we randomly choose some sketch and extrusion pairs from the current CAD construction sequence during a training step. We also randomly choose another construction sequence from the training dataset and select some sketch and extrusion pairs. We swap these sketch and extrusion pairs from these two selected construction sequences. 

\begin{figure}[!t]
\centering    
\includegraphics[width=.9\columnwidth]{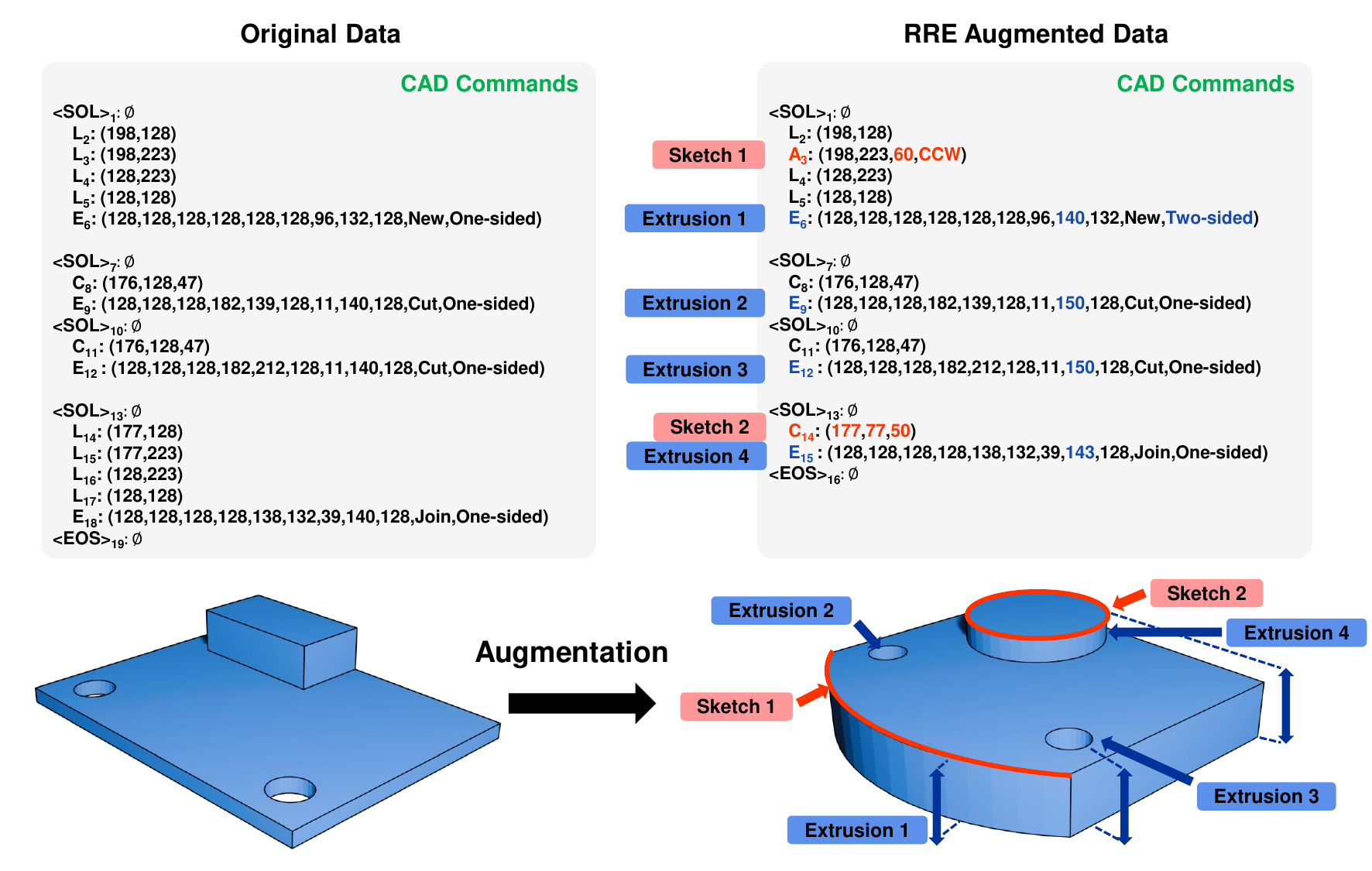}
   \caption{(a) Original CAD model example with construction sequences in the training dataset and (b) newly generated CAD model with construction sequences using the proposed RRE.} 
   \label{fig:fig4}
\end{figure}

Unlike previous approaches, the proposed RRE data augmentation can be applied to all construction sequences in the training dataset, regardless of the number of sketch and extrusion pairs. Figure \ref{fig:fig4} visualizes the construction sequences and their 3D shapes before and after applying the proposed RRE data augmentation method. As shown in the figure, the line command ($\text{L}_2$) is replaced with an arc command ($\text{A}_2$), introducing a new curve. Furthermore, changes in extrusion type (from One-sided to Two-sided in $\text{E}_1$) and extrusion distance (from 132 to 140 in $\text{E}_1$, from 128 to 132 in $\text{E}_1$, and from 140 to 150 in $\text{E}_2$, $\text{E}_3$) result in a new CAD model with a thicker base as shown in the figure. Furthermore, a shape in the form of a cuboid on the top in the original CAD model transforms into a completely new shape, a cylinder, in the new CAD model.

\subsection{ContrastCAD}
\subsubsection{Autoencoder}

The proposed autoencoder model takes a construction sequence $X_i=\left[C_1,\dots,C_N\right]$ consisting of a sequence of $N$ commands, $C_k$, as input. The encoder generates a latent vector, $z_i$, projected from $X_i$. Here, $X_i$ is the $i$-th element of a batch composed of m construction sequences. The decoder predicts a reconstructed construction sequence, $\hat{X}_i=\left[\hat{C}_1,\dots,\hat{C}_N\right]$, from this latent vector. The reconstruction loss ($l_{rec}$) between $X_i$ and $\hat{X}_i$ is then computed.

\paragraph{Encoder.} The encoder of ContrastCAD consists of a CAD embedding layer, a Transformer encoder, and an average pooling layer. First, the CAD embedding layer takes the input $X_i=\left[C_1,\dots,C_N\right]$ and outputs CAD embeddings $\left[e_1,\dots,e_N\right]$. Unlike natural language processing, $C_k$ is separated into $t_k$ that represents the type of command (i.e., {line, arc, circle, extrusion, $\langle\text{SOL}\rangle$, $\langle\text{EOS}\rangle$}) and $p_k\in\mathbb{R}^{1X16}$ that represents the parameters of the command. As explained in Section 3, there are 16 types of parameters (i.e., $p_k=\left[x,y,\theta,c,r,\alpha,\beta,\gamma,o_x,o_y,o_z,s,\delta_1,\delta_2,b,w\right]$). Therefore, embeddings need to be performed separately for $t_k$ and $p_k$. Let $d_E$ denotes the embedding dimension, the CAD embedding $e_k\in\mathbb{R}^{1×d_E)}$ for $C_k$ is calculated as follows by adding a learned positional encoding $e^{pos}$ to the embedding of $t_k$ and $p_k$: 
\begin{equation}
    e_k=t_k W^{cmd} + p_k W^{param} + e^{pos},
    \label{eq1}
\end{equation}
where $W^{cmd}\in\mathbb{R}^{1×d_E}$ and $W^{param}\in\mathbb{R}^{16×d_E}$ are weight matrices. 

Next, the Transformer encoder takes $\left[e_1,\dots,e_N\right]$ as input, performs self-attention and feed-forward computations, and outputs $\left[h_1,\dots,h_N\right]$. The Transformer encoder consists of $L$ stacked layers of self-attention and feed-forward layers, following the conventional Transformer encoder architecture. Finally, after passing through an average pooling layer, the latent vector $z_i$ ($i=1,\dots,m$) of $X_i$ is produced as output, where the dimension of $z_i$ is $d_E$.

\paragraph{Decoder.} The decoder of ContrastCAD consists of a Transformer decoder and a classifier layer. Similar to the Transformer encoder, the Transformer decoder comprises $L$ consecutive self-attention layers and feed-forward layers. The Transformer decoder takes a learned constant embedding as input and performs self-attention and feed-forward computations, attending to $z_i$. The output of the Transformer decoder is passed to the classifier layer, where it undergoes linear transformation and outputs softmax probabilities for the reconstructed construction sequence $\left[\hat{C}_1,\dots,\hat{C}_N\right]$ for $\left[C_1,\dots,C_N\right]$. Each $\hat{C}_k$ consists of the reconstructed command type $\hat{t}_k$ for $t_k$ and the reconstructed command parameter $\hat{p}_k$ for $p_k$.

\subsubsection{Contrastive Learning}
We propose to perform contrastive learning by augmenting embeddings using dropout techniques for CAD models. It forms positive pairs by applying two different dropout masks to the latent vectors. This approach is known to be effective in creating augmented views without data transformation \cite{simcse}. Our work is the first to apply this approach to CAD construction sequences. The contrastive learning portion of ContrastCAD comprises a projection layer and a masking layer based on dropout operations, as shown in Figure \ref{fig:fig3}(b). The projection layer, a simple linear layer, takes $z_i$ ($i=1,\dots,m$) as input and outputs $z_i^{proj}$ ($i=1,\dots,m$), projected into a new embedding space. The introduction of this projection layer is known to enhance the representation quality of $z_i$ and aid in contrastive learning \cite{simclr, simclr-v2}.

Then, $z_i^{proj}$ is inputted twice into the masking layer. The masking layer performs standard dropout operations with a probability of p for each of the two $z_i^{proj}$, using different dropout masks. Therefore, if we denote the outputs of the masking layer as $d_i$ and $d_j$ ($i=1,\dots,m$ and $j=m+1,\dots,2m$), $d_i$ and $d_j$ act as positive pairs in contrastive learning, having similar but different embedding values. In contrast, the outputs of other masking layers within the batch, excluding $d_i$, act as negative pairs with respect to $d_i$. By applying this embedding augmentation, we can better capture semantic information by generating augmented views for contrastive learning. They can be created for the same construction sequence without altering the shape of the CAD model represented by the input construction sequence.

\subsubsection{Training}
The autoencoder learns the representation of construction sequences in CAD models in an unsupervised manner and places latent vectors of similar CAD models closer together and those of dissimilar CAD models farther apart. Therefore, the proposed ContrastCAD trains the model using both reconstruction loss  ($l_{rec}$) and contrastive loss ($l_{cont}$). Here, $l_{rec}$ is utilized to learn the latent vectors of CAD construction sequences, while $l_{cont}$ is used for contrastive learning. Thus, the loss function of ContrastCAD is defined by combining $l_{rec}$ and $l_{cont}$ as follows:
\begin{equation}
    l_{ContrastCAD}=l_{rec}+\kappa l_{cont},
    \label{eq2}
\end{equation}
where $\kappa$ is a balancing hyperparameter between two terms. 

As each command in the construction sequence is divided into command type and parameters, it is necessary to learn representations for command types as well as command parameters. Therefore, $l_{rec}$ is calculated as the sum of cross-entropy loss between ($t_k$ and $\hat{t}_k$) and ($p_k$ and $\hat{p}_k$) and denoted as: 
\begin{equation}
    l_{rec}=CE(t_k,\hat{t}_k )+\lambda CE(p_k,\hat{p}_k),
    \label{eq3}
\end{equation}
where CE represents the standard cross-entropy loss, and $\lambda$ is as a balancing hyperparameter between the two terms that determines how much the cross-entropy loss for parameters is reflected in the overall $l_{rec}$.

In the contrastive learning portion of ContrastCAD, for a batch of size $m$, masking layers are applied twice to generate $2m$ augmented views as shown in Figure \ref{fig:fig3}(b). We reduce the distance in the latent space between positive pairs among the $2m$ augmented views and increase the distance between negative pairs. Therefore, it is necessary to compute the similarity between two output vector pairs (i.e., positive and negative pairs). We use cosine similarity to compute the similarity. Let the two output vectors of the masking layer be $u$ and $v$. Then, the similarity between $u$ and $v$, $\operatorname{SIM}(u, v)$, is calculated as follows: 
\begin{equation}
    \operatorname{SIM}(u, v)=\frac{u^{\top} v}{\|u\|\|v\|},
    \label{eq4}
\end{equation}
where $\|\cdot\|$ means L2-norm.

Therefore, a contrastive loss function, $l_{cont}$, is calculated to increase the similarity of positive pairs ($d_i$, $d_j$) while decreasing the similarity of negative pairs ($d_i$, $d_k$), which is computed as follows: 
\begin{equation}
    l_{\text {cont }}=-\log \frac{\exp \left(\operatorname{SIM}\left(d_i, d_j\right) / \tau\right)}{\sum_{k=1}^{2 m} \mathbb{I}[k \neq i] \exp \left(\operatorname{SIM}\left(d_i, d_k\right)\right)},
    \label{eq5}
\end{equation}
where $\mathbb{I}[k\neq i]\in\{1,0\}$ represents an indicator function that equals one iff k$\neq$i, and $\tau$ denotes the temperature. 

\subsection{CAD Generation}\label{subsec:cad generation}
When ContrastCAD is well-trained, it can automatically generate CAD models from latent vectors. In this study, we aim to perform CAD generation using latent-GAN. Unlike conventional GANs, latent-GAN learns over the latent space, enabling the generation of discrete data like construction sequences and capturing complex features of the latent space.

In our CAD generation method, the latent-GAN \cite{latentgan} first learns to generate new latent vectors from random noise. Specifically, the latent-GAN consists of a generator ($G$) and a discriminator ($D$). Both the generator and discriminator are simple multilayer perceptrons (MLPs), composed of four linear layers and non-linear activation functions. First, the generator takes random Gaussian noise $\epsilon\sim \mathcal{N}(0,I)$ as input to produce fake (generated) latent vectors $G(\epsilon)=\tilde{z}_i$ ($i=1,\dots,m$). The pre-trained encoder of ContrastCAD takes $X_i$ as input to generate real latent vectors $z_i$, which are used as ground truth for discriminator training. During CAD generation, all weights of ContrastCAD are frozen to prevent additional learning. The discriminator receives $z_i$ or $\tilde{z}_i$ as input and outputs logit scores to distinguish whether they are real or fake latent vectors. During training, the generator is trained to maximize the $l_g$ in order to fake the discriminator with the generated $\tilde{z}_i$ from $\epsilon$, which is denoted as: 
\begin{equation}
    l_g=D\left(G\left(\epsilon\right)\right).
    \label{eq6}
\end{equation}

In contrast, the discriminator is trained to minimize $l_d$, to effectively distinguish between $z_i$ and $\tilde{z}_i$. When we train the generator, the weights of the discriminator are frozen, and during discriminator training, the weights of the generator are frozen. Furthermore, for stable training, gradient penalties such as those used in \cite{wgan} are employed. The loss function for training latent-GAN is calculated as follows:
\begin{equation}
    l_d=D\left(G\left(\epsilon\right)\right)-D\left(z_i\right)+l_{gp},
    \label{eq7}
\end{equation}
where $l_{gp}$ is a gradient penalty and $\lambda$ is a balancing hyperparameter. 

Once the latentGAN is well-trained, we can obtain latent vectors $\tilde{z}_i$ generated from random noise $\epsilon$. Finally, by inputting these $\tilde{z}_i$ into the pre-trained decoder of ContrastCAD, we can generate construction sequences for new CAD models.

%%%%%%%%%%%%%%%%%%

\section{Experiments and Results}
\label{sec:Experiments and Results}
\subsection{Dataset}
We use the DeepCAD dataset for our experiments, which represents the CAD models using construction sequences and is widely used for training CAD models \cite{deepcad}. DeepCAD consists of a total of 178,238 construction sequences. We split the DeepCAD dataset into training/validation/testing sets with approximately a 90\%/5\%/5\% ratio. Consequently, the resulting data comprises 161,240 training samples, 8,946 validation samples, and 8,052 test samples.

\subsection{Implementation details}
All experiments were implemented using PyTorch and trained on an RTX A6000 GPU. Following \cite{deepcad}, we fixed $N$ to 60 for all construction sequences, and shorter sequences were padded with the $\langle\text{EOS}\rangle$ command. We set $d_E$ to 256, $L$ to 4, feed-forward dimension to 512, and the number of attention heads to 4. The dropout rate was set to 0.1. During training, we utilized the Adam optimizer \cite{adam} with an initial learning rate of 0.001, along with linear warmup for 2,000 steps and gradient clipping at 1.0. We trained for 1,000 epochs with a batch size ($m$) of 1,024. The $\lambda$ and $\kappa$ values for the loss function were set to 2, and $\tau$ was set to 0.07. The $\eta$ value in Equation \eqref{eq8} was set to 3.

\subsection{Effectiveness of RRE data augmentation method}
% Table 4
\begin{table}[]
\centering
\caption {Reconstruction performance of the Transformer-based autoencoder between the vanilla ContrastCAD without RRE and the ContrastCAD with RRE based on the command accuracy ($\text{ACC}_{\text{cmd}}$), parameter accuracy ($\text{ACC}_{\text{param}}$), invalid rate, and median Chamfer distance (CD).}
\label{tab:tab4}
\resizebox{.7\columnwidth}{!}{%
\begin{tabular}{lllll}
\Xhline{2\arrayrulewidth}
                   & $\text{ACC}_{\text{cmd}}$ & $\text{ACC}_{\text{param}}$ & \begin{tabular}[c]{@{}l@{}} Invalid \\ rate\end{tabular} & Median CD \\ \Xhline{2\arrayrulewidth}
ContrastCAD        & 99.67\%       & 98.49\%     & 3.23\%       & 0.831  \\
\begin{tabular}[c]{@{}l@{}} ContrastCAD \\ +RRE\end{tabular} & \textbf{99.81}\%       & \textbf{98.60}\%     & \textbf{2.55}\%       & \textbf{0.715}  \\ \Xhline{2\arrayrulewidth}
\end{tabular}%
}
\end{table}

\subsubsection{Baseline}
In this experiment, we test the effectiveness of the proposed RRE data augmentation method when it is combined with the ContrastCAD model. Specifically, we compare the reconstruction performance of the Transformer-based autoencoder on the DeepCAD test dataset between the vanilla ContrastCAD without RRE and the ContrastCAD with RRE. 

\subsubsection{Metrics}
In this experiment, performance evaluation was conducted using four metrics. First, to evaluate the reconstruction performance of ContrastCAD's output construction sequences compared to the input construction sequences, we measured \textit{command accuracy} ($\text{ACC}_{\text{cmd}}$) and \textit{parameter accuracy} ($\text{ACC}_{\text{param}}$) on the DeepCAD test dataset. $\text{ACC}_{\text{cmd}}$ measures the difference in command types between the input and reconstructed sequences, while $\text{ACC}_{\text{param}}$ measures the difference in command parameters between them. Each accuracy metric is calculated as follows:
\begin{equation}
\begin{aligned}
    \text{ACC}_{\text{cmd}} &=\frac{1}{N} \sum_{k=1}^N \mathbb{I}\left[t_k=\hat{t}_k\right], \\
    \text{ACC}_{\text{param}}   &=\frac{1}{T} \sum_{k=1}^N \sum_{l=1}^{\left|\hat{p}_k\right|} \mathbb{I}\left[\left|p_{k, l}-\hat{p}_{k, l}\right|<\eta\right] \mathbb{I}\left[t_k=\hat{t}_k\right],
\end{aligned}\label{eq8}
\end{equation}
where $T =\sum_{k=1}^N \mathbb{I}\left[t_k=\hat{t}_k\right]\left|p_k\right|$. Here, $T$ represents the number of correctly predicted parameters, $|p_k|$ denotes the number of parameter types (=16), and $\eta$ is the tolerance for parameter accuracy \cite{deepcad}.

We also measured the median \textit{Chamfer distance} (CD) of point clouds to evaluate the reconstruction performance of the 3D geometry of the CAD models. We evaluated the median CD by sampling 2,000 point clouds from the 3D shape generated from both the input sequence and the reconstructed sequence. Finally, to assess the extent of invalid topology in the 3D CAD models generated from ContrastCAD's output construction sequences, we measured the \textit{invalid rate}. Here, we considered validity based on whether the CAD model could be created by the CAD kernel and transformed into a point cloud among the entire reconstructed construction sequences \cite{deepcad}.

\subsubsection{Results}
Table \ref{tab:tab4} compares the reconstruction performance of the output construction sequences and the resulting CAD model's 3D geometry. We observe that the ContrastCAD with RRE demonstrates higher reconstruction performance in all four metrics compared to the vanilla ContrastCAD without RRE. In particular, we observe a 13.9\% improvement in the median CD of ContrastCAD. The experimental results show that augmenting data with RRE enhances the learning capability of ContrastCAD and improves its reconstruction performance. The main reason for the significant improvement in $\text{ACC}_{\text{param}}$ observed in the experimental results is that the arc command, which previously had poor performance owing to being the scarcest within the dataset, experienced enhanced learning performance of the model through proposed RRE data augmentation. As a result, there was a substantial enhancement in $\text{ACC}_{\text{param}}$ for the arc command.

\begin{figure}[!t]
\centering    
\includegraphics[width=.7\columnwidth]{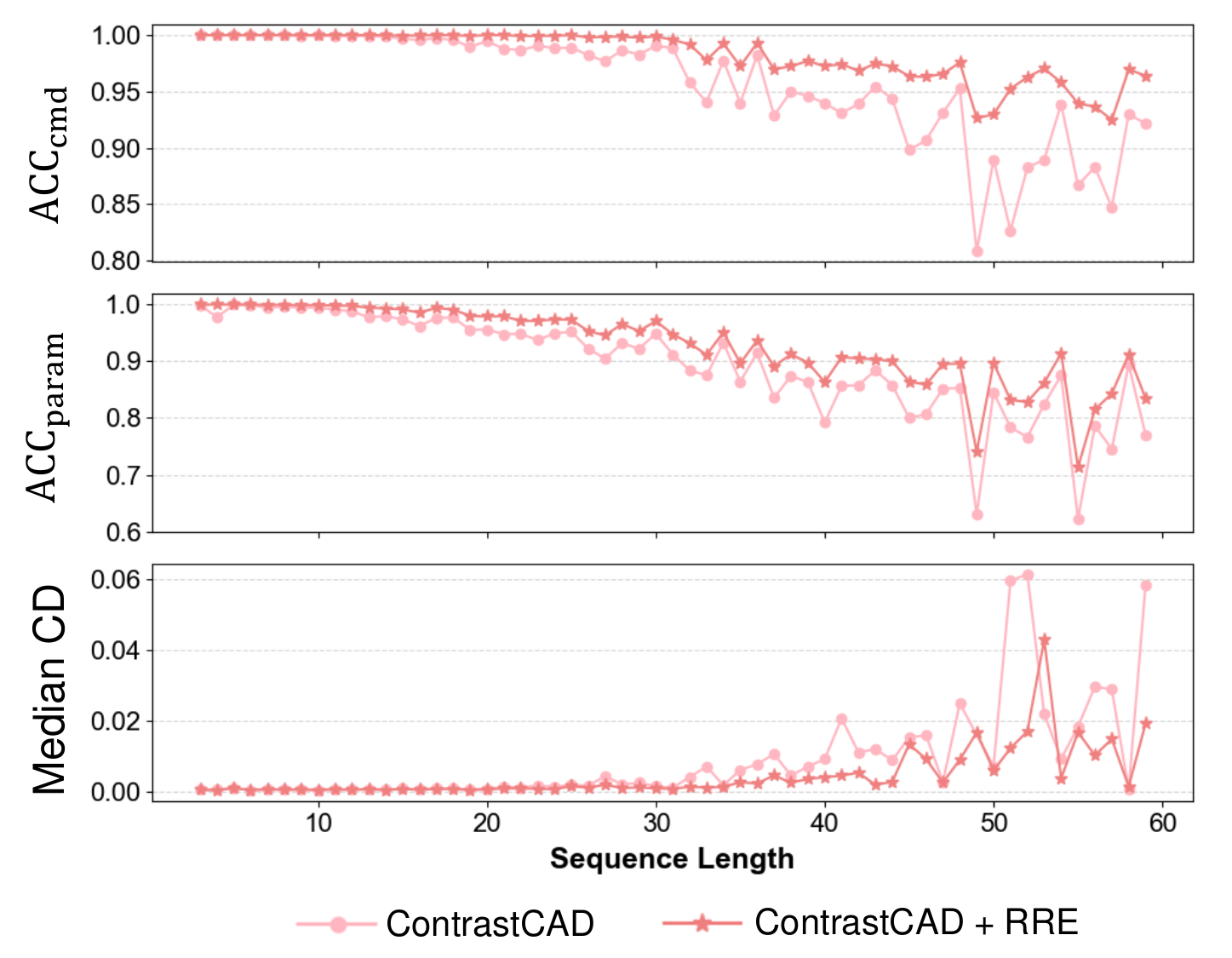}
    \caption{Command accuracy ($\text{ACC}_{\text{cmd}}$), parameter accuracy $\text{ACC}_{\text{param}}$, and median CD with respect to the lengths of construction sequences between the vanilla ContrastCAD without RRE and the ContrastCAD with RRE.} 
    \label{fig:fig5}
\end{figure}

Figure \ref{fig:fig5} illustrates the command accuracy ($\text{ACC}_{\text{cmd}}$), parameter accuracy ($\text{ACC}_{\text{param}}$), and median CD depending on the length of construction sequences. We observe that ContrastCAD with RRE exhibits higher accuracy across all sequence lengths compared to vanilla ContrastCAD without RRE. One of the notable results is that the proposed RRE data augmentation significantly improves the learning of long construction sequences. As CAD models with longer construction sequences mostly represent complex 3D shapes, it is crucial for models to learn these long construction sequences well. The experimental results demonstrate that the effectiveness of the proposed RRE data augmentation becomes more evident as the length of the construction sequence increases. We observe that the performance gap between ContrastCAD with RRE and vanilla ContrastCAD without RRE increases as the length of the construction sequence increases, especially for construction sequences that exceed a length of 38. For example, for construction sequences with a length 59, which is the longest, ContrastCAD with RRE exhibited improvements of 4.24\% in $\text{ACC}_{\text{cmd}}$, 6.39\% in $\text{ACC}_{\text{param}}$, and 29.70\% in median CD compared to vanilla ContrastCAD without RRE.

\subsection{Evaluation of representation learning performance}

% Table 5
\begin{table}[t!]
\centering
\caption {Similarity results between the original construction sequences and the newly generated construction sequences with permutation changes measured by the cosine similarity ($\operatorname{SIM}$) and euclidean distance (ED).}
\label{tab:tab5}
\resizebox{.4\columnwidth}{!}{%
\begin{tabular}{lll}
\Xhline{2\arrayrulewidth}
            & $\operatorname{SIM}$            & ED             \\ \Xhline{2\arrayrulewidth}
DeepCAD \cite{deepcad}  & 0.913          & 0.386          \\
ContrastCAD & \textbf{0.922} & \textbf{0.362} \\ \Xhline{2\arrayrulewidth}
\end{tabular}
}
\end{table}

\subsubsection{Baseline}
We evaluate the representation learning performance of the proposed ContrastCAD by comparing the baseline model, DeepCAD, which is composed solely of a Transformer-based autoencoder without any contrastive learning. To purely evaluate the performance improvement of the proposed contrastive learning, both ContrastCAD and the baseline model, DeepCAD, were used for experiments without employing any data augmentation methods. 

\subsubsection{Metrics}
In this experiment, we evaluate the representation learning performance of ContrastCAD by using four evaluation metrics to assess the latent space of ContrastCAD and DeepCAD. We evaluated how well the latent vectors of similar CAD models are clustered in the latent space. As the original DeepCAD dataset does not have class labels, we performed K-Means clustering on the latent space by assigning cluster labels to the latent vectors. Subsequently, using the assigned cluster labels, we measured the \textit{silhouette coefficient} (SC) and \textit{sum squared error} (SSE) as evaluation metrics for each latent space. Let $x_i$ be a $i$-th data point in a set of size n in the dataset $X$. Let $a_i$ denote the average distance between $x_i$ and the other points in its cluster, and $b_i$ denote the average distance between $x_i$ and points in other clusters. The SC is defined as follows:
\begin{equation}
    \text{SC}=\frac{b_i-a_i}{\max \left(a_i, b_i\right)}.
    \label{eq9}
\end{equation}

The SC takes values between $-1$ and $1$, where a value closer to one indicates better clustering. Next, the SSE is calculated as the sum of the squared distances between the data points and the centroids: 
\begin{equation}
    \text{SSE}=\sum_{i=1}^n\left(x_i-\bar{x}_i\right)^2,
    \label{eq10}
\end{equation}
where $\bar{x}_i$ is the centroid of $x_i$. Lower SSE values indicates better clustering. 

To measure the similarity between latent vectors between two CAD models, we use the SIM and Euclidean distance (ED). Let the latent vectors for two construction sequences be $u$ and $v$. Then, ED is calculated as follows:
\begin{equation}
    \text{ED}(u, v)=\|u-v\|,
    \label{eq11}
\end{equation}
where $\|\cdot\|$ means L2-norm. A lower ED value indicates better similarity, with a minimum value of zero. 

\subsubsection{Clustering results using a K-means algorithm}
\begin{figure}[!t]
\centering    
\includegraphics[width=.7\columnwidth]{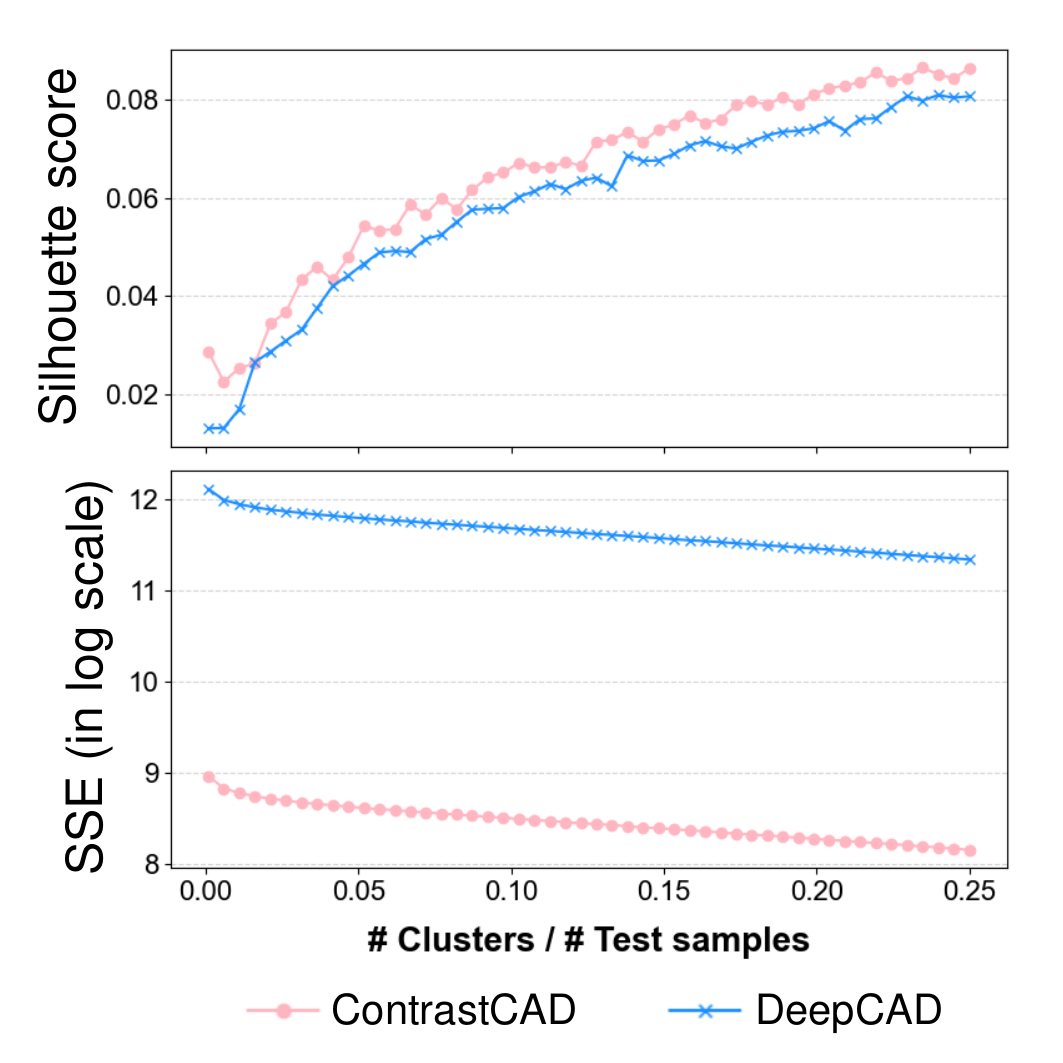}
    \caption{Clustering results using the K-means algorithm between ContrastCAD and DeepCAD.} 
    \label{fig:fig6}
\end{figure}

Figure \ref{fig:fig6} shows the results by measuring SC and SSE as the number of clusters ($K$) increases when K-means clustering method is employed. We observe that the ContrastCAD model shows much better clustering results with higher SC and lower SSE values compared to the baseline model. ContrastCAD shows the best results when the “number of clusters / number of test samples” is 0.25. ContrastCAD exhibits a 6.55\% increase in SC and a 28.02\% decrease in SSE compared to the baseline model. Our results show that ContrastCAD brings similar shapes of CAD models closer together in the latent spaces, ultimately resulting in a more highly clustered representation space for similar CAD models.

\subsubsection{Results on robustness of the model with respect to permutation changes}
In this experiment, we evaluate the robustness of ContrastCAD with respect to permutation changes of construction sequences. We expect ContrastCAD to be able to learn to position the latent vectors of construction sequences of CAD models that have identical shapes but different permutations close to each other in the latent space.

We generate a new construction sequence by performing permutation changes on the original construction sequence of CAD models. Note that the newly generated construction sequences have the same shapes as the original construction sequence. We generate new construction sequences by changing the permutations of original construction sequences that include three sketch and extrusion patterns, P1: ($\langle\text{SOL}\rangle$, L, L, L, L, E), P2: ($\langle\text{SOL}\rangle$, L, L, L, L, L, L, E), and P3: ($\langle\text{SOL}\rangle$, C, $\langle\text{SOL}\rangle$, C, E), which appear most frequently in the test dataset. When we change the permutations of the original construction sequence for the patterns P1 and P2, we maintain the orientation of the original construction consequence but shift the commands. For example, if the original construction includes a sketch and extrusion pattern that has ($\text{L}_1$, $\text{L}_2$, $\text{L}_3$, $\text{L}_4$), a new construction sequence randomly includes one of these three sketch and extrusion patterns: ($\text{L}_2$, $\text{L}_3$, $\text{L}_4$, $\text{L}_1$), ($\text{L}_3$, $\text{L}_4$, $\text{L}_1$, $\text{L}_2$), and ($\text{L}_4$, $\text{L}_1$, $\text{L}_2$, $\text{L}_3$), with the same orientation. For the third sketch and extrusion pattern P3, we generate a new construction sequence by switching the two circle commands in the construction sequence. Finally, we measure the similarity between the original construction sequence and the newly generated construction sequence in the latent space.  

In Table \ref{tab:tab5}, we compare the similarity results of latent vectors between the original construction sequence and new construction sequences with a different permutation. From the results in the table, we observe that the ContrastCAD model exhibits higher similarity compared to DeepCAD when measured using $\operatorname{SIM}$ and ED. Specifically, $\operatorname{SIM}$ and ED have been improved by 0.98\% and 6.22\%, respectively, in ContrastCAD compared to DeepCAD. Our experimental results show that ContrastCAD can position CAD models having the same shape but different permutations closer in the latent space compared to DeepCAD.

\subsubsection{Results on similarity among similar CAD models}
\begin{figure}[!t]
\centering    
    \includegraphics[width=.9\columnwidth]{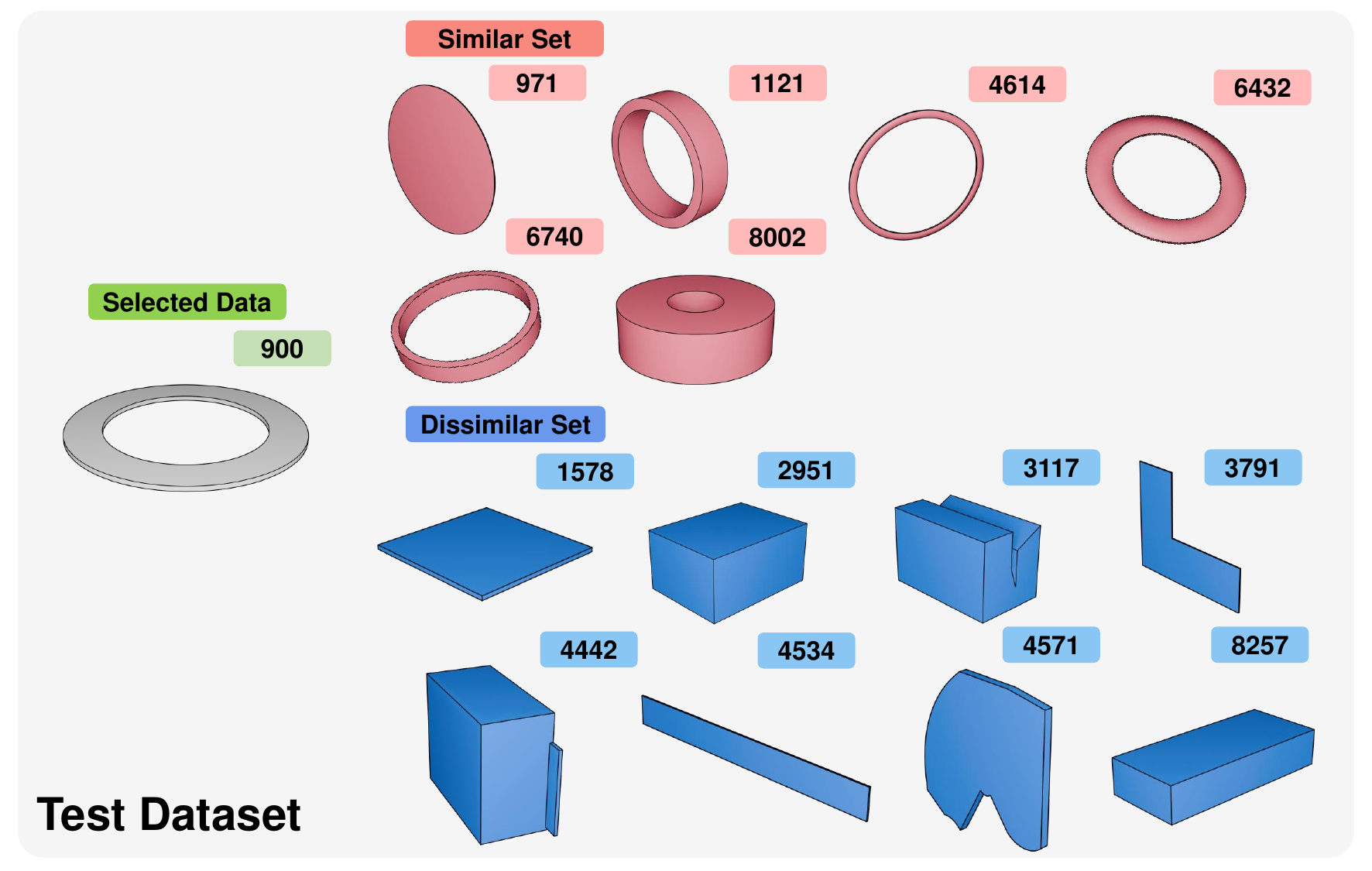}
    \caption{Various CAD models that have similar shapes (similar set) and dissimilar shapes (dissimilar set) with the CAD model (index 900) that are used to test similarity in the latent space.} 
    \label{fig:fig7}
\end{figure}

In this experiment, we evaluate how closely CAD modes that have similar shapes and dissimilar shapes are positioned in the latent space after training ContrastCAD. As illustrated in Figure \ref{fig:fig7}, we select five CAD models (indices 971, 1121, 6432, 6740, and 8002) from the test dataset that are visually similar to a specific CAD model (index 900) and define them as a \textit{similar set}. We also select eight other CAD models (index 1578, 2951, 3117, 3791, 4442, 4534, 4571, and 8257) from the test dataset that visually differ from the CAD model with index 900 and define them as the \textit{dissimilar set}.

\begin{figure*}[ht]
  \centering
  \includegraphics[width=.8\columnwidth]{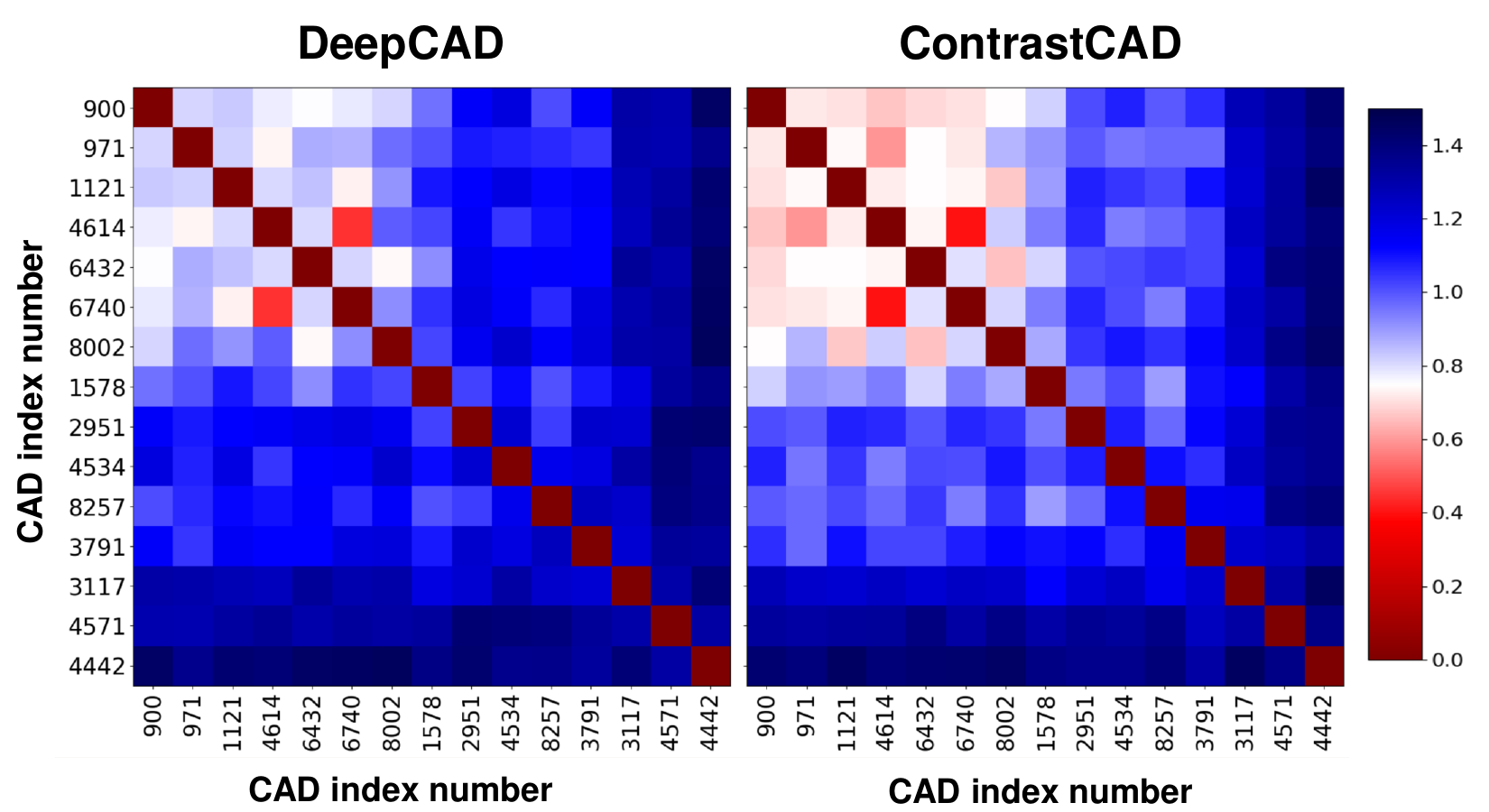}
  \caption{Similarity matrix evaluated by the Euclidean distances among the latent vectors of the compared 15 CAD models.}
  \label{fig:fig8}
\end{figure*}

In Figure \ref{fig:fig8}, we present the similarity matrix evaluated by the Euclidean distances among the latent vectors of the 15 compared CAD models used in the experiment. We observe that the similarity among CAD models belonging to the similar set have smaller ED values compared to those among CAD models belonging to the dissimilar set in the latent space. We also observe that as shown in the top-left portion of Figure \ref{fig:fig8}, the proposed ContrastCAD demonstrates smaller Euclidean distances between CAD models belonging to the similar set in the latent space compared to DeepCAD. Specifically, when calculating the average Euclidean distance in the latent space between a CAD model with index 900 and other CAD models within the similar set, the value for DeepCAD is 0.8, whereas for ContrastCAD it is 0.67, which means that similar CAD models are more clustered for the ContrastCAD compared with the DeepCAD. ContrastCAD is able to learn to position CAD models with similar shapes closer together in the latent space compared to DeepCAD.

\subsection{CAD Generation Performance}

% Table 6
\begin{table}[t!]
\centering
\caption {Evaluation results for Validity, Uniqueness, COV, JSD, and MMD of the generated 3D CAD models}
\label{tab:tab6}
\resizebox{.7\columnwidth}{!}{%
\begin{tabular}{llllll}
\Xhline{2\arrayrulewidth}
                   & Valid $\uparrow$ & Unique $\uparrow$  & COV $\uparrow$ & JSD $\downarrow$ & MMD $\downarrow$ \\ 
                   \Xhline{2\arrayrulewidth}
DeepCAD \cite{deepcad}            & 86.25\%               & 89.84\%                & 78.33\%             & 3.92           & \textbf{1.43}  \\
ContrastCAD        & 87.30\%               & 90.65\%                & 78.50\%             & 3.90           & 1.44           \\
\begin{tabular}[c]{@{}l@{}}ContrastCAD\\+RRE\end{tabular} & \textbf{87.42}\%      & \textbf{90.86}\%       & \textbf{78.93}\%    & \textbf{3.67}  & 1.44           \\ \Xhline{2\arrayrulewidth}
\end{tabular}%
}
\end{table}

\subsubsection{Baseline}
In this experiment, we compared the CAD generation performance of ContrastCAD with RRE, ContrastCAD without RRE, and DeepCAD to evaluate the CAD model generation performance. Both methods utilized the latent GAN approach described in Section \ref{subsec:cad generation} during CAD model generation.

\subsubsection{Metrics}
We randomly generated 9,000 sample CAD models and measured the \textit{Validity} and \textit{Uniqueness} of the generated 3D CAD models. Validity refers to the proportion of construction sequences within the entire generated sequence that can produce CAD models by CAD kernels and can be converted into point clouds \cite{deepcad}. Uniqueness represents the proportion of samples within the generated set that are not duplicated. For Validity as well as Uniqueness metrics, higher values indicate better performance. 

To evaluate the 3D CAD models generated from ContrastCAD's latent vectors, we measure the difference in the geometric domain between the test samples and the generated samples. We extracted point clouds from the generated 3D CAD models and measured \textit{Coverage} (COV), \textit{Jensen–Shannon divergence} (JSD), and \textit{Minimum Matching Distance} (MMD) \cite{latentgan, deepcad}. COV represents the percentage of ground truth samples that match generated samples, JSD measures the similarity between the distributions of the generated set and the ground truth set, and MMD denotes the minimum matching distance between generated samples and their nearest neighbors in the ground truth set. Here, the ground truth samples refer to the samples from the DeepCAD test dataset. Higher COV, lower JSD, and lower MMD indicate better performance. 

\subsubsection{Results}
Table \ref{tab:tab6} presents the evaluation results for Validity, Uniqueness, COV, JSD, and MMD of the generated 3D CAD models. As observed from the table, the CAD generation performance of ContrastCAD with RRE outperforms other compared models. Specifically, ContrastCAD with RRE achieves enhancements of 1.17\% and 1.02\%, respectively, when measured by the Validity and Uniqueness compared to those with DeepCAD.

From these results, we observe that the proposed ContrastCAD-based models effectively learn the distribution of CAD models in the training dataset and successfully generate new CAD models from random noise. This indicates that ContrastCAD-based models produce fewer invalid topologies compared to DeepCAD and can generate various 3D CAD models without duplication. The latent vectors trained by the ContrastCAD-based models also lead to substantial performance improvements when generating new 3D CAD models compared to DeepCAD.

Figure \ref{fig:fig9} illustrates examples of 3D CAD models generated using ContrastCAD with RRE. As evident in the illustration, the proposed model successfully generates diverse and complex 3D CAD models.

\begin{figure}[!t]
\centering    
\includegraphics[width=.9\columnwidth]{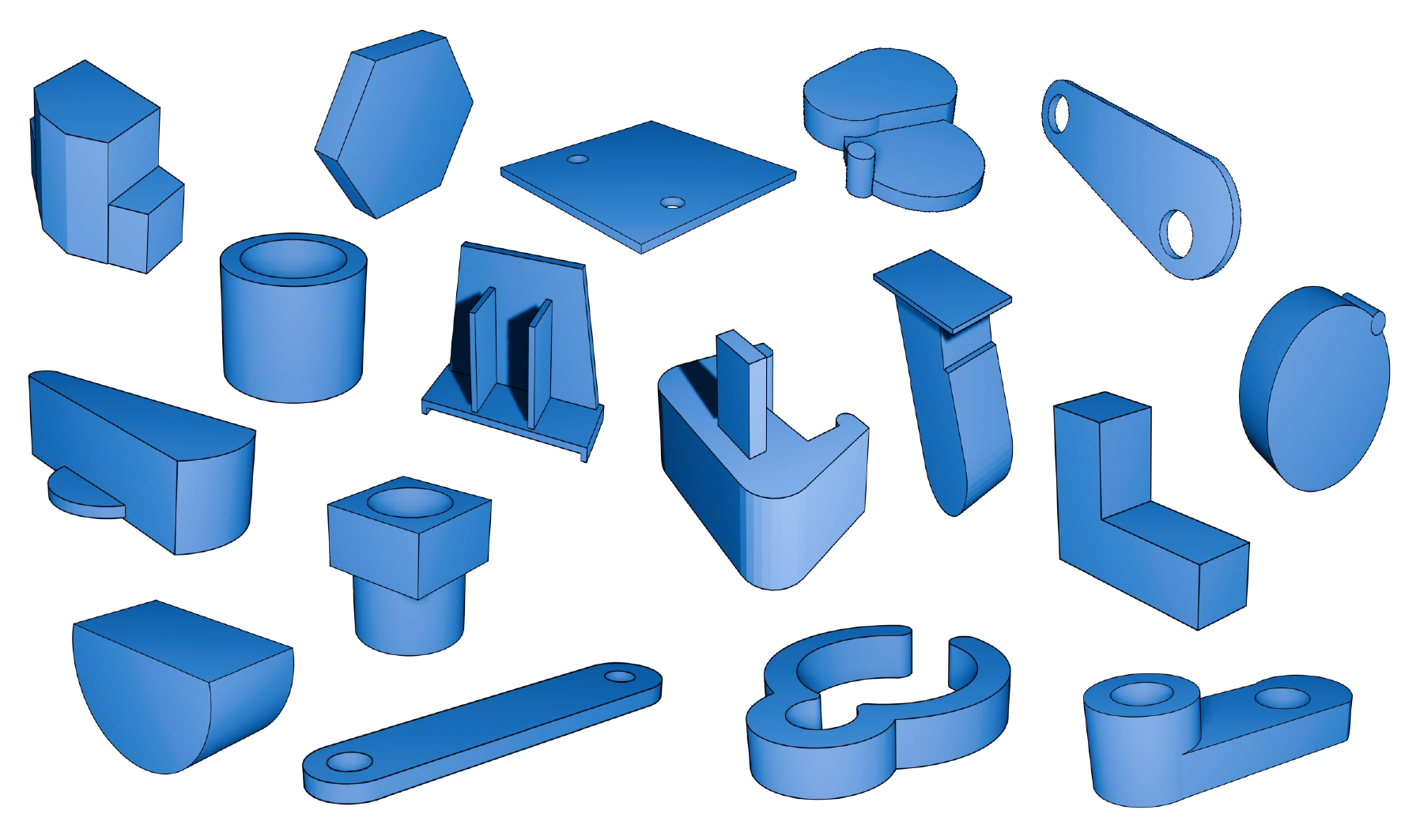}
   \caption{Examples of 3D CAD models generated using the ContrastCAD model with RRE} 
   \label{fig:fig1}
\end{figure}

\section{Conclusion}
\label{sec:conclusion}
In this paper, we propose a novel contrastive learning-based approach for learning CAD models expressed in CAD construction sequences. The proposed ContrastCAD performs contrastive learning by augmenting embeddings, allowing for better reflection of the semantic information of CAD models. We also introduce a new CAD data augmentation method called RRE data augmentation for enhancing the learning capability of the model with an imbalanced CAD training dataset. Our experimental results show that the new RRE data augmentation method significantly improves the reconstruction accuracy of the autoencoder. In particular, it considerably improves the long sequence learning problem by improving the reconstruction accuracy of the input CAD sequences. The proposed ContrastCAD achieves better representation learning and therefore, CAD models with similar shapes are positioned closely in the latent space. We also observe that the proposed ContrastCAD model is robust to the permutation changes of construction sequences.

CAD models can be represented in various forms of multimodal data such as point cloud, mesh, and sketch. In future work, we plan to further research multimodal learning methods that can simultaneously learn from multiple forms of multimodal data. There is still a limitation where only random shapes of CAD models are generated instead of the shapes desired by the user during CAD model generation. We plan to conduct research on conditional generation of CAD models, where users can input text or images to generate the desired CAD model.


\begin{thebibliography}{10}

\bibitem{point2geo}
Lingxiao Li, Minhyuk Sung, Anastasia Dubrovina, Li~Yi, and Leonidas~J Guibas.
\newblock Supervised fitting of geometric primitives to 3d point clouds.
\newblock In {\em Proc. IEEE/CVF Conf. Comput. Vis. Pattern Recognit.}, pages
  2652--2660, 2019.

\bibitem{parsenet}
Gopal Sharma, Difan Liu, Subhransu Maji, Evangelos Kalogerakis, Siddhartha
  Chaudhuri, and Radom{\'\i}r M{\v{e}}ch.
\newblock Parsenet: A parametric surface fitting network for 3d point clouds.
\newblock In {\em Proc. Eur. Conf. Comput. Vis.}, pages 261--276. Springer,
  2020.

\bibitem{pienet}
Xiaogang Wang, Yuelang Xu, Kai Xu, Andrea Tagliasacchi, Bin Zhou, Ali
  Mahdavi-Amiri, and Hao Zhang.
\newblock Pie-net: Parametric inference of point cloud edges.
\newblock In {\em Proc. Adv. Neural Inf. Process. Syst.}, volume~33, pages
  20167--20178, 2020.

\bibitem{inferring-cad}
Xianghao Xu, Wenzhe Peng, Chin-Yi Cheng, Karl~DD Willis, and Daniel Ritchie.
\newblock Inferring cad modeling sequences using zone graphs.
\newblock In {\em Proc. IEEE/CVF Conf. Comput. Vis. Pattern Recognit.}, pages
  6062--6070, 2021.

\bibitem{deepcad}
Rundi Wu, Chang Xiao, and Changxi Zheng.
\newblock Deepcad: A deep generative network for computer-aided design models.
\newblock In {\em Proc. IEEE/CVF Int. Conf. Comput. Vis.}, pages 6772--6782,
  2021.

\bibitem{skexgen}
Xiang Xu, Karl~DD Willis, Joseph~G Lambourne, Chin-Yi Cheng, Pradeep~Kumar
  Jayaraman, and Yasutaka Furukawa.
\newblock Skexgen: Autoregressive generation of cad construction sequences with
  disentangled codebooks.
\newblock In {\em Proc. Int. Conf. Mach. Learn.}, pages 24698--24724. PMLR,
  2022.

\bibitem{hiernet}
Xiang Xu, Pradeep~Kumar Jayaraman, Joseph~G. Lambourne, Karl~D.D. Willis, and
  Yasutaka Furukawa.
\newblock Hierarchical neural coding for controllable cad model generation.
\newblock In {\em Proc. Int. Conf. Mach. Learn.}, ICML'23. JMLR.org, 2023.

\bibitem{voxel2cad}
Joseph~George Lambourne, Karl Willis, Pradeep~Kumar Jayaraman, Longfei Zhang,
  Aditya Sanghi, and Kamal~Rahimi Malekshan.
\newblock Reconstructing editable prismatic cad from rounded voxel models.
\newblock In {\em Proc. SIGGRAPH Asia}, SA ’22. ACM, November 2022.

\bibitem{structurenet}
Kaichun Mo, Paul Guerrero, Li~Yi, Hao Su, Peter Wonka, Niloy~J. Mitra, and
  Leonidas~J. Guibas.
\newblock Structurenet: hierarchical graph networks for 3d shape generation.
\newblock {\em ACM Trans. Graph.}, 38(6), nov 2019.

\bibitem{shapeassembly}
R.~Kenny Jones, Theresa Barton, Xianghao Xu, Kai Wang, Ellen Jiang, Paul
  Guerrero, Niloy Mitra, and Daniel Ritchie.
\newblock Shapeassembly: Learning to generate programs for 3d shape structure
  synthesis.
\newblock {\em ACM Trans. Graph.}, 39(6):Article 234, 2020.

\bibitem{uvnet}
Pradeep~Kumar Jayaraman, Aditya Sanghi, Joseph~G. Lambourne, Karl D.~D. Willis,
  Thomas Davies, Hooman Shayani, and Nigel Morris.
\newblock Uv-net: Learning from boundary representations.
\newblock {\em arXiv:2006.10211}, 2021.

\bibitem{brepnet}
Joseph~G Lambourne, Karl~DD Willis, Pradeep~Kumar Jayaraman, Aditya Sanghi,
  Peter Meltzer, and Hooman Shayani.
\newblock Brepnet: A topological message passing system for solid models.
\newblock In {\em Proc. IEEE/CVF Conf. Comput. Vis. Pattern Recognit.}, pages
  12773--12782, 2021.

\bibitem{complexgen}
Haoxiang Guo, Shilin Liu, Hao Pan, Yang Liu, Xin Tong, and Baining Guo.
\newblock Complexgen: Cad reconstruction by b-rep chain complex generation.
\newblock {\em ACM Trans. Graph.}, 41(4):1--18, 2022.

\bibitem{csgnet}
Gopal Sharma, Rishabh Goyal, Difan Liu, Evangelos Kalogerakis, and Subhransu
  Maji.
\newblock Csgnet: Neural shape parser for constructive solid geometry.
\newblock In {\em Proc. IEEE/CVF Conf. Comput. Vis. Pattern Recognit.}, June
  2018.

\bibitem{ucsgnet}
Kacper Kania, Maciej Zieba, and Tomasz Kajdanowicz.
\newblock Ucsg-net-unsupervised discovering of constructive solid geometry
  tree.
\newblock In {\em Proc. Adv. Neural Inf. Process. Syst.}, volume~33, pages
  8776--8786, 2020.

\bibitem{d2csg}
Fenggen Yu, Qimin Chen, Maham Tanveer, Ali Mahdavi~Amiri, and Hao Zhang.
\newblock D\^{}2csg: Unsupervised learning of compact csg trees with dual
  complements and dropouts.
\newblock In {\em Proc. Adv. Neural Inf. Process. Syst.}, volume~36, 2024.

\bibitem{sketch2cad}
Changjian Li, Hao Pan, Adrien Bousseau, and Niloy~J Mitra.
\newblock Sketch2cad: Sequential cad modeling by sketching in context.
\newblock {\em ACM Trans. Graph.}, 39(6):1--14, 2020.

\bibitem{vitruvion}
Ari Seff, Wenda Zhou, Nick Richardson, and Ryan~P Adams.
\newblock Vitruvion: A generative model of parametric cad sketches.
\newblock In {\em Proc. Int. Conf. Learn. Represent.}, 2021.

\bibitem{free2cad}
Changjian Li, Hao Pan, Adrien Bousseau, and Niloy~J. Mitra.
\newblock Free2cad: Parsing freehand drawings into cad commands.
\newblock {\em ACM Trans. Graph.}, 41(4):93:1--93:16, 2022.

\bibitem{cad-sketches}
Hao~Pan Yuezhi~Yang.
\newblock Discovering design concepts for cad sketches.
\newblock In {\em Proc. Adv. Neural Inf. Process. Syst.}, volume~35, 2022.

\bibitem{abcdataset}
Sebastian Koch, Albert Matveev, Zhongshi Jiang, Francis Williams, Alexey
  Artemov, Evgeny Burnaev, Marc Alexa, Denis Zorin, and Daniele Panozzo.
\newblock Abc: A big cad model dataset for geometric deep learning.
\newblock In {\em Proc. IEEE/CVF Conf. Comput. Vis. Pattern Recognit.}, pages
  9601--9611, 2019.

\bibitem{simclr}
Ting Chen, Simon Kornblith, Mohammad Norouzi, and Geoffrey Hinton.
\newblock A simple framework for contrastive learning of visual
  representations.
\newblock {\em arXiv:2002.05709}, 2020.

\bibitem{simclr-v2}
Ting Chen, Simon Kornblith, Kevin Swersky, Mohammad Norouzi, and Geoffrey
  Hinton.
\newblock Big self-supervised models are strong semi-supervised learners.
\newblock {\em arXiv:2006.10029}, 2020.

\bibitem{moco}
Kaiming He, Haoqi Fan, Yuxin Wu, Saining Xie, and Ross Girshick.
\newblock Momentum contrast for unsupervised visual representation learning.
\newblock In {\em Proc. IEEE/CVF Conf. Comput. Vis. Pattern Recognit.}, pages
  9729--9738, 2020.

\bibitem{byol}
Jean-Bastien Grill, Florian Strub, Florent Altch{\'e}, Corentin Tallec, Pierre
  Richemond, Elena Buchatskaya, Carl Doersch, Bernardo Avila~Pires, Zhaohan
  Guo, Mohammad Gheshlaghi~Azar, et~al.
\newblock Bootstrap your own latent-a new approach to self-supervised learning.
\newblock In {\em Proc. Adv. Neural Inf. Process. Syst.}, volume~33, pages
  21271--21284, 2020.

\bibitem{siamsiam}
Xinlei Chen and Kaiming He.
\newblock Exploring simple siamese representation learning.
\newblock In {\em Proc. IEEE/CVF Conf. Comput. Vis. Pattern Recognit.}, pages
  15750--15758, 2021.

\bibitem{simcse}
Tianyu Gao, Xingcheng Yao, and Danqi Chen.
\newblock {SimCSE}: Simple contrastive learning of sentence embeddings.
\newblock In {\em Proc. Empirical Methods Natural Lang. Process.}, 2021.

\bibitem{diffcse}
Yung-Sung Chuang, Rumen Dangovski, Hongyin Luo, Yang Zhang, Shiyu Chang, Marin
  Soljacic, Shang-Wen Li, Wen-tau Yih, Yoon Kim, and James Glass.
\newblock {DiffCSE}: Difference-based contrastive learning for sentence
  embeddings.
\newblock In {\em Proc. Annu. Conf. North Amer. Chapter Assoc. Comput.
  Linguistics}, 2022.

\bibitem{multicad}
Weijian Ma, Minyang Xu, Xueyang Li, and Xiangdong Zhou.
\newblock Multicad: Contrastive representation learning for multi-modal 3d
  computer-aided design models.
\newblock In {\em Proc. ACM Int. Conf. Inf. Knowl. Manage.}, pages 1766--1776,
  2023.

\bibitem{latentgan}
Panos Achlioptas, Olga Diamanti, Ioannis Mitliagkas, and Leonidas Guibas.
\newblock Learning representations and generative models for 3d point clouds.
\newblock In {\em Proc. Int. Conf. Mach. Learn.}, pages 40--49. PMLR, 2018.

\bibitem{wgan}
Martin Arjovsky, Soumith Chintala, and L{\'e}on Bottou.
\newblock Wasserstein generative adversarial networks.
\newblock In {\em Proc. Int. Conf. Mach. Learn.}, pages 214--223. PMLR, 2017.

\bibitem{adam}
Diederik Kingma and Jimmy Ba.
\newblock Adam: A method for stochastic optimization.
\newblock In {\em Proc. Int. Conf. Learn. Represent.}, San Diego, CA, USA, May
  2015.

\end{thebibliography}
\end{document}